\documentclass[journal]{IEEEtran}

\usepackage{cite}
\usepackage{microtype}
\usepackage{graphicx}
\usepackage{booktabs}
\usepackage{tikz}
\usetikzlibrary{positioning}
\usepackage{amsmath}
\usepackage{amssymb}
\usepackage{mathtools}
\usepackage{amsthm}
\usepackage[hidelinks]{hyperref}
\usepackage[capitalize,noabbrev]{cleveref}

\theoremstyle{plain}

\theoremstyle{definition}

\theoremstyle{remark}

\begin{document}
\title{The Predictive--Optimization Gap in GuanDan Self-Play}

\author{Qicheng~Zou, Biao~Yuan, Siyu~Song, Wentao~Liu, and~Xiaokun~Wang%
\thanks{Q.~Zou is with the University of California San Diego, La Jolla, CA, USA.}%
\thanks{B.~Yuan, S.~Song, and W.~Liu are with Chabiyue (Shanghai) Information Technology Co., Ltd., Shanghai, China.}%
\thanks{X.~Wang is with Chabiyue (Shanghai) Information Technology Co., Ltd., Shanghai, China (corresponding author, e-mail: chubbyue@163.com).}%
}

\maketitle
\thispagestyle{plain}
\pagestyle{plain}
\bstctlcite{BSTcontrol}

\begin{abstract}
Outcome-predictive signals are often treated as promising optimization targets in reinforcement learning, but predictive validity and optimization utility are different claims.
We expose this distinction in GuanDan, a multi-round imperfect-information game in which the implication of a round outcome depends on the public match state.
Held-out analysis shows that state--gain interactions provide incremental predictive value beyond a constant signed-gain effect, establishing that match progress is predictively meaningful.
We then use changes in a frozen match-win predictor, $\Delta\Phi$, as a controlled round-level supervision intervention in a matched DanZero-style Monte Carlo self-play protocol.
The intervention can succeed in an individual lineage: a selected progress-augmented checkpoint substantially outperforms its matched Baseline on a locked evaluation.
Yet the method-level result reverses this impression: none of five preregistered replication seeds is positive, the pooled ten-seed interval includes zero, and across-iteration effects repeatedly change sign without a common late-stage advantage.
These results identify a predictive--optimization gap: under the studied learner, a signal can describe outcome-relevant progress while failing to provide a reliably beneficial optimization direction.
\end{abstract}


\section{Introduction}
\label{sec:introduction}

Outcome prediction and policy optimization are often coupled by an appealing intuition: if an intermediate signal better predicts terminal success, then exposing that signal to the learner should improve credit assignment.
The first implication is statistical, whereas the second is algorithmic.
A signal is \emph{predictively meaningful} when it contains held-out information about eventual outcomes beyond a simpler alternative; it is \emph{optimization reliable} when using it produces a consistently beneficial policy effect across stochastic training runs.
The broader predict-then-optimize literature already shows that minimizing predictive loss and maximizing downstream decision quality are different objectives \cite{smart_predict_optimize,decision_focused}.
Prediction alone does not guarantee optimization reliability because policy learning also depends on credit granularity, function approximation, finite-sample updates, and the feedback loop between a changing policy and its self-generated data; related performative-learning work formalizes how deployed decisions can change the distribution on which later learning occurs \cite{performative_prediction,performative_rl}.
This paper isolates that usually implicit gap.

GuanDan provides a natural test bed for separating these two claims.
It is a four-player, team-based, imperfect-information card game in which a complete match spans multiple rounds as the public level state of each team evolves \cite{openguandan}.
DouZero established large-scale self-play deep Monte Carlo learning for DouDizhu, while DanZero-family work transferred this foundation to GuanDan and DanZero+ further refined a pretrained DMC policy using PPO \cite{DouZero,DanZero,DanZeroPlus}.
Unlike settings in which the episode boundary and the principal competitive objective coincide, GuanDan has a temporal hierarchy: actions determine round outcomes, and a sequence of rounds determines the match winner.
Consequently, the same one-level gain can have a different implication for eventual victory early in a match than in a close, near-terminal state.

This hierarchy lets us ask two questions that are related but not equivalent.
First, does the predictive relationship between a fixed signed round gain and eventual victory depend on the pre-round public state?
Second, if that dependence is exposed explicitly as round-level supervision, does it yield a reproducible improvement in self-play optimization?
The first question concerns held-out predictive content; the second concerns the distribution of policy effects over independent training lineages.
Separating them requires both a predictive test and method-level replication rather than inferring optimization value from predictor quality or from one successful checkpoint.

We construct a controlled probe of this predictive--optimization gap.
Following DanZero-style Monte Carlo practice \cite{DouZero,DanZero}, the Baseline combines signed round gain with the eventual match outcome and broadcasts that label to every executed action in the round.
We train and freeze a public-state match-win predictor $\Phi$, then define match progress as the change in predicted winning probability across a completed round, $\Delta\Phi$.
The progress-augmented target is
\begin{equation}
G^{\mathrm{prog}}
=
G^{\mathrm{base}}
+
\beta\Delta\Phi,
\label{eq:progress_compact}
\end{equation}
where $G^{\mathrm{base}}$ is the baseline round-level label and $\beta$ controls the contribution of match progress.
We use this augmentation as a diagnostic intervention, not as an assumption that a more predictive label must be a better reward.
Because both arms share the same initialization, architecture, learner, budget, and evaluation protocol, their difference tests whether predictive match context is converted into reliable policy improvement under this specific round-broadcast, executed-action regression mechanism.

The two stages of evidence lead to a counterintuitive result.
On held-out matches, adding pre-round state--gain interactions improves log loss over a constant-gain specification, and the frozen predictor produces semantically consistent, state-dependent progress estimates.
Thus, the signal is predictively meaningful by the criterion defined above.
At optimization time, however, the selected seed-1 Progress-10 checkpoint beats its matched Baseline by $0.143$ MeanDelta on a locked 1,000-pair evaluation, while none of the five preregistered replication seeds has a positive effect and the pooled ten-seed interval includes zero.
Across-iteration comparisons repeatedly cross zero and show no common late-stage Progress advantage.
The central finding is therefore not that match progress is useless, nor merely that one reward variant failed; it is that a verified predictive signal can support a strong selected policy without yielding a reliable optimization effect.

This distinction changes how intermediate supervision should be validated.
Predictor accuracy establishes what a signal describes under an observed distribution, while across-run optimization evidence establishes whether a learner can use it consistently as its policy and data distribution co-evolve.
A successful checkpoint demonstrates possibility, but it cannot substitute for replication over training randomness.
For outcome-potential supervision, these are separate empirical obligations.
Accordingly, we do not claim to be the first to observe that prediction and decision objectives can differ.
Our contribution is a controlled self-play case study that tests predictor validity, selected-checkpoint efficacy, and method-level optimization reliability within one matched intervention, then shows that the three conclusions separate.

Our contributions are summarized as follows:
\begin{itemize}
    \item We construct a two-stage self-play diagnostic that separately evaluates predictor validity, selection-conditioned checkpoint efficacy, and optimization reliability across independent training runs.
    \item We establish that GuanDan round gains have state-dependent predictive meaning using held-out state--gain interactions and a calibrated, frozen public-state outcome model.
    \item We document a predictive--optimization gap under round-broadcast Monte Carlo regression: selected-checkpoint success coexists with failed method-level replication, heterogeneous ten-seed effects, and sign-changing training-horizon trajectories.
\end{itemize}

\section{Background and Related Work}
\label{sec:related_work}

\subsection{Self-Play in Card Games and Training Reliability}

Imperfect-information card games provide challenging environments for reinforcement learning because agents must act under hidden information, large and variable action spaces, and mixed competitive--cooperative interaction.
DouZero and DanZero-family systems established large-scale self-play learning for DouDizhu and GuanDan \cite{DouZero,DanZero,DanZeroPlus}, while OpenGuanDan and related work such as Big~2 further demonstrate the applicability of this paradigm \cite{openguandan,big2_selfplay}.
Most existing card-game RL systems primarily address policy learning, representation, action-space complexity, self-play scalability, or imperfect-information handling.
Our work retains an online self-play setting while focusing on how match-progress information is exposed to round-level learning, rather than proposing a new policy architecture or imperfect-information solver.

Reliable evaluation in deep RL requires uncertainty over independent training runs, because extensive evaluation of one frozen policy does not measure how retraining changes the learned policy \cite{deep_rl_matters,statistical_precipice}.
Self-play adds an endogenous source of variation: each policy helps generate the opponents and state distribution used for subsequent learning.
In nontransitive games, self-play can exhibit strategic cycles rather than monotonic improvement \cite{open_ended_selfplay}; we do not claim that this specific mechanism is identified in GuanDan, but it motivates treating lineage-dependent trajectories as a method-level object rather than as evaluation noise.
Our protocol accordingly separates fixed-checkpoint uncertainty from variability across matched self-play training runs.

\subsection{Predictive Validity and Decision Utility}

Predict-then-optimize and decision-focused learning distinguish statistical prediction loss from the quality of decisions induced downstream.
Smart Predict-then-Optimize defines loss through decision error, while decision-focused learning differentiates through an optimization layer, because standard predictive metrics need not be reliable proxies for downstream utility \cite{smart_predict_optimize,decision_focused}.
These works establish the general prediction--decision mismatch that motivates our study; our novelty is not the mismatch itself.
Instead, we test how it appears when a validated predictor-derived signal is inserted into an online self-play regression target and evaluated at three levels: predictor validity, selection-conditioned checkpoint efficacy, and across-run optimization reliability.

Feedback makes this distinction sharper in adaptive systems.
Performative prediction studies settings in which deployed predictions alter the distribution of later observations, and performative RL extends this idea to policies that affect subsequent rewards or dynamics \cite{performative_prediction,performative_rl}.
Our environment and frozen predictor differ from those formal models, but the common structural point is that freezing a scoring function does not freeze the data distribution induced by decisions.
In GuanDan self-play, the progress term changes policy updates, and the updated policy generates later training matches; held-out predictive improvement therefore supplies no monotonic guarantee for the resulting optimization path.

\subsection{Outcome-Potential and Intermediate Supervision}

Temporal credit assignment concerns how delayed outcomes should inform earlier decisions \cite{temporal_credit_survey}.
Reward-redistribution methods such as RUDDER and randomized return decomposition move delayed trajectory returns toward more informative intermediate events, often seeking return-equivalent rearrangements of the same total return \cite{rudder,rrd}.
Value transport likewise carries long-horizon outcome information to earlier decisions \cite{value_transport}.
These approaches motivate intermediate supervision, but they generally operate over primitive timesteps, actions, or trajectory events.

A complementary line constructs intermediate supervision from predictions of eventual task success.
Outcome-potential methods evaluate intermediate states by how favorable they appear with respect to the final outcome and use changes in that evaluation as denser feedback; recent examples in long-horizon language-model agent settings include SIOP, TRACE, and TIPS \cite{siop,trace,tips}.
Although these methods operate in different domains, they motivate converting outcome predictions into intermediate supervision.
We study the same conversion at completed-round boundaries, but test predictive content and optimization reliability separately rather than treating predictor quality as evidence that the learner will benefit.

\subsection{Policy Guarantees and Causal Credit}

Differences between state potentials are closely related in form to potential-based reward shaping (PBRS).
In the classical formulation, an additional transition reward $F(s,a,s')=\gamma\Phi(s')-\Phi(s)$ preserves the optimal policy under the corresponding Markov decision process assumptions \cite{ng1999rewardshaping}; related results extend invariance properties to multiplayer stochastic games under appropriate conditions \cite{lu2011stochasticgames}.
Our round-boundary signal is only \emph{potential-difference inspired}: it augments an executed-action Monte Carlo regression label and is not a policy-preserving reward transformation.

Predictive progress is also distinct from causal credit.
Counterfactual credit-assignment methods estimate how outcomes would differ under alternative actions or contributions \cite{implicit_multiagent_credit,counterfactual_credit}, whereas $\Delta\Phi_k$ describes the observed change in public-state win probability after a realized round.
It does not isolate the marginal contribution of an executed action, and every action in the round receives the same label.
Moreover, adding $\beta\Delta\Phi_k$ does not preserve trajectory return as in return-equivalent redistribution.
The absence of policy-invariance, return-equivalence, and causal-credit guarantees makes optimization benefit an empirical question even after predictive meaning has been established.

\section{Diagnostic Intervention}
\label{sec:method}

We construct a diagnostic intervention that connects a validated predictive signal to policy optimization without presuming that the connection will be beneficial.
The design has three parts: a Baseline round-level target, a frozen public-state predictor whose held-out value can be tested independently, and a matched augmentation that changes only the supervision target supplied to the self-play learner.
This separation lets us ask whether a predictively meaningful state--gain signal becomes an optimization-reliable training signal under a fixed learning protocol.

\subsection{Problem and Round-Level Targets}
\label{sec:problem_targets}
\label{sec:problem_formulation}
\label{sec:baseline_return}
\label{sec:delta_phi}
\label{sec:progress_target}

We consider GuanDan as a sequence of rounds embedded in a longer match.
Let $x_k$ denote the public match state before round $k+1$, and let $\tau_k=\{(s_{k,t},a_{k,t})\}_{t=1}^{T_k}$ denote the within-round trajectory of executed actions.
The game therefore contains a natural temporal hierarchy, $\text{action}\rightarrow\text{round}\rightarrow\text{match}$.
Our work focuses on the second boundary: how information about the eventual match outcome can be assigned to individual rounds, without attempting within-round action-level attribution.

For a team-relative sample, let $y_{\mathrm{match}}\in\{-1,+1\}$ denote the final match outcome, and let $r^{\mathrm{rank}}_k$ denote the signed level gain from round $k$.
In our implementation, the nonzero round outcomes take magnitudes corresponding to one-, two-, or three-level gains.
Following DanZero-style Monte Carlo practice, the baseline round-level supervision target is
\begin{equation}
G^{\mathrm{base}}_{k,t}
=
r^{\mathrm{rank}}_k
+
y_{\mathrm{match}},
\label{eq:baseline_reward}
\end{equation}
Every executed action within round $k$ receives the same label, and $y_{\mathrm{match}}$ is broadcast to all rounds after the match ends.
We call $G^{\mathrm{base}}_{k,t}$ a \emph{round-level Monte Carlo supervision target}: it is the regression label used in executed-action learning, not a standard return-to-go obtained by accumulating subsequent round rewards.
For a fixed realized pair $(r^{\mathrm{rank}}_k,y_{\mathrm{match}})$, the scalar label does not vary with the pre-round public match state.
For example, two one-level gains that end in the same match outcome receive identical baseline labels even if one occurs early in a match and the other occurs near a terminal level configuration.
Because the network observes $s$ and is trained against $y_{\mathrm{match}}$, it can in principle learn $\mathbb{E}[G^{\mathrm{base}}\mid s,a]$ that already varies with match context.
An explicit progress term therefore does not supply missing state access; it tests whether presenting the same outcome-relevant dependence as a round-local teacher produces more favorable finite-sample updates than recovering it implicitly from terminal outcomes.

Let $\Phi_\psi(x)$ denote a frozen public-state estimate of match-winning probability.
For a completed round $k$, the match-progress signal is
\begin{equation}
\Delta\Phi_k
=
\Phi_\psi(x_{k+1})-\Phi_\psi(x_k).
\label{eq:delta_phi}
\end{equation}
A positive value indicates that the public match state became more favorable to the evaluated team, whereas a negative value indicates the opposite.
Unlike the raw level gain, $\Delta\Phi_k$ is explicitly state dependent: the same rank result may correspond to different changes in estimated winning probability at different stages of the match.
Eq.~\eqref{eq:delta_phi} should not be interpreted as causal attribution; it is a predictive teacher signal under the data-generating policy distribution.

The progress-augmented target expands Eq.~\eqref{eq:progress_compact} as
\begin{equation}
G^{\mathrm{prog}}_{k,t}
=
r^{\mathrm{rank}}_k
+
y_{\mathrm{match}}
+
\beta\Delta\Phi_k,
\label{eq:progress_reward}
\end{equation}
where $\beta$ controls the contribution of match progress and every executed action in round $k$ again shares the same round label.
Thus the diagnostic augments a target that already contains local rank feedback and the broadcast final match outcome.
The initial optimization probe instantiates two preregistered scales, Progress-10 and Progress-20, with arm-specific target scaling constants locked before policy training from a fixed reference dataset.

\subsection{Frozen Predictor}
\label{sec:phi}

We train an independently fitted public-state match-win predictor $\Phi_\psi(x)=P_\psi(y_{\mathrm{match}}=+1\mid x)$, where $x$ contains only information publicly available at a round boundary.
Throughout, $\Phi_\psi$ denotes this frozen outcome model---a policy-distribution-conditional win-probability teacher---rather than a policy-training value function or online critic.
It is trained on complete-match trajectories, calibrated, and then frozen before all reward-training experiments; it is never updated from subsequent self-play trajectories.
Freezing separates match-progress estimation from policy optimization and prevents the reward annotation itself from becoming a moving target as the self-play policy changes.

The public-state restriction excludes hidden cards, private observations, future round outcomes, and policy-internal representations unavailable to the acting agent.
Train/validation/test partitions are constructed at the complete-match level so that different rounds from the same match cannot appear in different splits.
The predictor uses a 27-dimensional public-boundary representation encoding team levels and gap, stage indicators, and other publicly known match-state variables.
We fit an $\ell_2$-regularized logistic model and apply scalar Platt calibration on the validation split, which is also used for predictor development; there is no separate calibration partition.
All reported predictor metrics are evaluated once on untouched test matches.
The frozen artifact stores the feature transformation, coefficients, calibration parameters, clipping rule, and a cryptographic identifier.
Opponent predictions are defined by construction as $\Phi_{\mathrm{opp}}(x)=1-\Phi_{\mathrm{team}}(x)$, so the corresponding progress signals are antisymmetric.

\subsection{Self-Play Optimization}
\label{sec:dmc_training}

Policy learning follows an iterative online self-play procedure rather than offline reinforcement learning.
At iteration $i$, the current actor $\pi_i$ generates new complete matches; after each match ends, the frozen predictor supplies $\Delta\Phi_k$ and the corresponding round-level supervision targets are materialized.
The learner minimizes the executed-action regression objective
\begin{equation}
\mathcal{L}_{\mathrm{MC}}(\theta)
=
\frac{1}{N}
\sum_{(s,a,G)}
\left(
Q_\theta(s,a)-G
\right)^2,
\label{eq:mc_loss}
\end{equation}
where $G$ is either Eq.~\eqref{eq:baseline_reward} or Eq.~\eqref{eq:progress_reward}.
Only the executed action is supervised by its realized round-level label; no counterfactual target is available for unexecuted legal actions.

Replacing $G^{\mathrm{base}}$ by $G^{\mathrm{prog}}=G^{\mathrm{base}}+\beta\Delta\Phi_k$ shifts every within-round residual by $-\beta\Delta\Phi_k$.
Because $\Delta\Phi_k$ varies with public match context even when $r^{\mathrm{rank}}_k$ is held fixed, samples that share the same rank term but occur in different stages receive systematically different residual corrections, changing the finite-sample update direction at round boundaries relative to recovering the same dependence only through the delayed and widely shared $y_{\mathrm{match}}$.
After locked arm-specific scaling, the progress component is calibrated to a controlled fraction of reference target RMS, so the intervention changes residual structure and teacher locality rather than merely rescaling the objective; we do not claim that this necessarily improves signal-to-noise ratio in every stratum.

The Q-network contains 4,961,793 parameters.
A residual MLP with LayerNorm maps the 638-dimensional state input to width 512 using two residual blocks; a GRU maps the 94-dimensional action-history sequence to width 256; and a residual MLP maps the 89-dimensional candidate-action input to width 256.
Their representations are fused by two width-768 residual blocks followed by a shared linear Q head, with dropout 0.05.
Both reward arms use the identical architecture and initialization.

\begin{figure}[!t]
\centering
\begin{tikzpicture}[
    font=\footnotesize,
    >=latex,
    box/.style={
        draw,
        rounded corners=2pt,
        align=center,
        text width=3.55cm,
        minimum height=0.72cm,
        inner sep=3pt
    },
    sidebox/.style={
        draw,
        densely dashed,
        rounded corners=2pt,
        align=center,
        text width=1.55cm,
        inner sep=2.5pt
    },
    arrow/.style={
        ->,
        thick
    }
]
    \node (pi) [box] {Current policy $\pi_i$};
    \node (collect) [box, below=0.4cm of pi]
        {Complete-match collection\\[-1pt]
         {\scriptsize new complete matches}};
    \node (annotate) [box, below=0.55cm of collect]
        {Round-boundary annotation\\[-1pt]
         {\scriptsize after match ends}\\[-1pt]
         {\scriptsize $\Delta\Phi=\Phi(x_{\mathrm{after}})-\Phi(x_{\mathrm{before}})$}\\[-1pt]
         {\scriptsize broadcast to actions in the round}};
    \node (train) [box, below=0.4cm of annotate]
        {Round-label regression training\\[-1pt]
         {\scriptsize one epoch; optimizer reset}};
    \node (pnext) [box, below=0.4cm of train]
        {Updated policy $\pi_{i+1}$};

    \node (phi) [sidebox, left=0.28cm of annotate]
        {Frozen $\Phi$\\all iters};

    \draw [arrow] (pi) -- (collect);
    \draw [arrow] (collect) -- (annotate);
    \draw [arrow] (annotate) -- (train);
    \draw [arrow] (train) -- (pnext);
    \draw [arrow] (phi) -- (annotate);

    \draw [arrow]
        (pnext.east) -- ++(0.48,0) |- (collect.east)
        node[pos=0.22, right, font=\scriptsize, align=left]
        {next\\iteration};
\end{tikzpicture}
\caption{One online collect--annotate--train iteration.
Each iteration first generates new complete matches with the current policy $\pi_i$.
After a match ends, the frozen predictor $\Phi$ annotates each round boundary with $\Delta\Phi=\Phi(x_{\mathrm{after}})-\Phi(x_{\mathrm{before}})$, and the resulting round-level supervision target is broadcast to every executed action within that round.
These round-level labels then supervise one epoch of policy training with a fresh optimizer state, yielding $\pi_{i+1}$ for the next collection round.
The predictor $\Phi$ remains frozen across all policy-training iterations and is not updated with the policy.}
\label{fig:flowchart}
\end{figure}
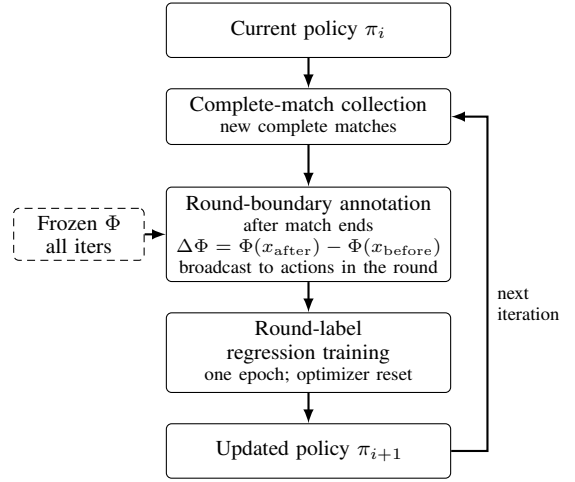

The data lifecycle for iteration $i$ is shown in Figure~\ref{fig:flowchart}.

Each iteration collects 250 newly generated complete matches and performs one epoch over the resulting executed-action samples with batch size 2048 and learning rate $5\times10^{-5}$.
Training runs for eight iterations.
$\epsilon$-greedy exploration decreases from 0.05 to 0.01, and optimizer state is reset between iterations.
Temporary reinforcement-learning trajectories are discarded after training.

\section{Testing Predictive Meaning and Optimization Reliability}
\label{sec:experiments}

\subsection{Complete-Match Dataset}
\label{sec:match_dataset}

A dedicated complete-match collector was constructed to support both the match-outcome predictor and the reward experiment.
The dataset used in the formal experiments was generated with the frozen DMC self-play Parent and contains 2{,}000 complete matches with zero aborted matches.
The collection contains 24{,}636 rounds, 2{,}398{,}084 executed decisions, and 49{,}272 team-relative round-boundary samples.
Matches contain 12.3 rounds on average, with a 95th percentile of 17 rounds and a maximum of 24 rounds.

\begin{table}[!t]
\caption{Complete-match dataset partitions. Boundary rows include both team-relative views.}
\label{tab:data_split}
\centering
\begin{small}
\begin{sc}
\begin{tabular}{lrr}
\toprule
Split & Matches & Boundaries \\
\midrule
Train & 1,411 & 34,818 \\
Val.\ (Platt) & 199 & 4,886 \\
Test & 390 & 9,568 \\
\midrule
Total & 2,000 & 49,272 \\
\bottomrule
\end{tabular}
\end{sc}
\end{small}
\end{table}

Dataset partitions are defined by \texttt{match\_id}, and no match contributes boundaries to more than one partition.
Each boundary record includes the public states before and after the round, signed level gain, final match outcome, and team-relative match information.

\subsection{Stage I: Testing Predictive Meaning of State--Gain Interactions}
\label{sec:r2}

Before constructing the progress reward, we measure whether the association between a fixed signed level gain and eventual victory changes with the pre-round public match state.
A trivial comparison between level gain alone and full match state would be insufficient, since variables such as the current team levels naturally predict the final winner.
A positive answer indicates that an explicit state--gain interaction carries additional predictive content that a round-local teacher can expose more directly than a shared terminal label alone.

We compare two nested predictive specifications. The constant-effect model is
\begin{equation}
\operatorname{logit}
P(y_{\mathrm{match}}=1\mid x_k,\Delta_k)
=
f(x_k)
+
c\Delta_k,
\label{eq:r2_m0}
\end{equation}
where $\Delta_k$ is the signed round-level gain and the coefficient $c$ is shared across all match states.
The state-dependent model is
\begin{equation}
\operatorname{logit}
P(y_{\mathrm{match}}=1\mid x_k,\Delta_k)
=
f(x_k)
+
c\Delta_k
+
h(x_k,\Delta_k),
\label{eq:r2_m1}
\end{equation}
where $h$ contains preregistered interactions between round gain and pre-round match-state variables.
Only variables available before the round are included in $x_k$; variables determined by the round outcome are excluded to avoid post-treatment leakage.

We evaluate the two models on held-out complete matches using log loss and Brier score as primary metrics, with AUROC and calibration diagnostics as secondary measures.
Confidence intervals are obtained by match-level resampling, keeping team-relative views from the same match in the same cluster.
For a distribution-level audit, we condition on realized signed level gain $\Delta_k\in\{-3,-2,-1,+1,+2,+3\}$ and stratify by pre-round level gap (leading / close / trailing), with Near-A status retained as an overlapping Boolean slice.
Stage contrasts use match-clustered percentile bootstrap intervals.
This is an observational stratification of realized outcomes rather than a same-deal intervention.
A meaningful held-out improvement of Eq.~\eqref{eq:r2_m1} over Eq.~\eqref{eq:r2_m0} establishes predictive meaning for the state--gain interaction, but does not by itself establish optimization utility.

\subsection{Constructing the Frozen Predictive Probe}
\label{sec:r3}

We train $\Phi_\psi$ on the complete-match train split with Platt calibration on validation, without access to the final test matches.
Predictor quality is evaluated by log loss, Brier score, reliability, stage-stratified calibration, team-view consistency, and later out-of-support rates under self-play.
Model and calibration parameters are then frozen under a fixed artifact hash; out-of-support states use a deterministic fallback and are logged.

\subsection{Stage II: Matched Self-Play Optimization Probe}
\label{sec:reward_protocol}
\label{sec:parent_init}
\label{sec:reward_conditions}
\label{sec:r4_training}
\label{sec:selection}

All formal reward-training arms are initialized from the same frozen DMC self-play Parent.
This policy was originally initialized through supervised learning on human data and then improved in a separate pre-experiment self-play stage using the conventional round-win objective; the resulting iteration-4 deployment champion was frozen before the reward experiment.
Human-data replay is disabled during formal training ($w_{\mathrm{BC}}=w_{\mathrm{humanMC}}=0$), so human supervision enters only through the frozen initialization lineage.

The initial optimization-probe experiment contains one Baseline arm and two preregistered progress scales.
Progress-10 and Progress-20 target auxiliary progress components equal to approximately 10\% and 20\% of the frozen reference target RMS, respectively, with locked constants $c_{\mathrm{base}}=2.556$, $c_{\mathrm{P10}}=2.747$, $c_{\mathrm{P20}}=2.948$, $\beta_{10}=2.002$, and $\beta_{20}=4.005$.
All three arms use the same Parent, network, DMC learner, exploration schedule, collection budget, training duration, and evaluation seeds; after training begins, each arm induces its own on-policy state distribution.
We evaluate the reward under a DanZero-style DMC learner rather than the full DanZero+ PPO system.
Progress-10 is selected only after the initial five-seed optimization comparison; Progress-20 does not receive a final-holdout evaluation.

\paragraph{Training-protocol safeguards.}
Preliminary Monte Carlo runs showed that bounded-target mean-squared regression on a preference-oriented scoring head can compress output scale and degrade action ranking.
Formal arms therefore share the same Parent and training protocol, with the complete reward target as the principal manipulated treatment; checkpoints from preliminary harmful runs are excluded from subsequent self-play collection.

Formal reward training proceeds in iterative collect--train cycles under the protocol of Section~\ref{sec:dmc_training}.
Scientific training lineage is maintained separately from deployment promotion.
The initial scale-selection experiment uses five independent training seeds per arm and eight iterations of 250 completed matches per seed, for $3\times5\times8\times250=30{,}000$ completed training matches.
After Progress-10 was frozen, a preregistered method-level replication trained five additional matched Baseline and Progress-10 lineages (seeds 6--10) under the same Parent, learner, hyperparameters, budget, and scientific-checkpoint selection rule, adding $2\times5\times8\times250=20{,}000$ completed matches.
The combined training evidence therefore covers ten Baseline--Progress-10 seed pairs and 50{,}000 completed matches, while Progress-20 remains confined to the initial scale-selection stage.

Checkpoint selection uses 40 paired seeds to reject clearly harmful candidates and 200 disjoint paired seeds for confirmation; the scientific checkpoint is chosen by complete-match paired performance.
The final evaluation set remains untouched during training, hyperparameter selection, and checkpoint confirmation.
After Progress-10 is selected, the corresponding scientific checkpoint is evaluated on a locked final holdout of 1{,}000 paired seeds against both the Parent and the matched Baseline.

\subsection{Paired Team-Swap Evaluation}
\label{sec:paired_eval}

Playing strength is evaluated using complete matches under a paired team-swap protocol.
For each evaluation seed, two matches are played under the same controlled conditions while swapping which team is controlled by the candidate policy.
Let $m_{i,A},m_{i,B}\in\{-1,+1\}$ denote the candidate's outcomes in the two team assignments.
The paired outcome is
\begin{equation}
\bar d_i
=
\frac{m_{i,A}+m_{i,B}}{2}
\in\{-1,0,+1\},
\label{eq:paired_outcome}
\end{equation}
so $\bar d_i=+1$ denotes a double win, $\bar d_i=-1$ a double loss, and $\bar d_i=0$ a split pair.
Our primary paired metric is
\begin{equation}
\operatorname{MeanDelta}
=
\frac{1}{n}
\sum_{i=1}^{n}
\bar d_i
=
\frac{N_{+}-N_{-}}{n},
\label{eq:mean_delta}
\end{equation}
where $N_{+}$ and $N_{-}$ are the numbers of double-win and double-loss pairs.
When every individual match has a binary win/loss outcome, $\operatorname{MeanDelta}=2\operatorname{WinRate}-1$.
Inference is performed over paired seeds because the two team-swapped matches are deliberately coupled by the evaluation design.

\subsection{Criterion for Optimization Reliability}
\label{sec:seed_robustness}

We define optimization reliability at the level of independent training runs rather than evaluation games from one selected checkpoint.
This follows deep-RL evaluation guidance that run-to-run variability and finite-run uncertainty must be reported explicitly when comparing learning procedures \cite{deep_rl_matters,statistical_precipice}.
Increasing the number of evaluation games tightens uncertainty for a fixed checkpoint, but does not estimate the variability of policies produced by rerunning training.
For training seed $s$, let $\Delta^{\mathrm{prog}}_s$ and $\Delta^{\mathrm{base}}_s$ denote the paired performance of the progress and baseline policies under matched evaluation conditions, and define
\begin{equation}
D_s
=
\Delta^{\mathrm{prog}}_s
-
\Delta^{\mathrm{base}}_s.
\label{eq:seed_effect}
\end{equation}
We report the individual $D_s$ values, their mean and median, the fraction of positive training seeds, and an uncertainty interval over training seeds.
The original seeds 1--5 use the 200-pair confirmation estimates.
The preregistered seeds 6--10 use a fresh 1{,}000-pair common-random-number evaluation pool, and their Student-$t$ interval is the primary method-level replication inference; pooling seeds 1--10 is a prespecified secondary summary.

Following the original confirmation analysis for Progress-10, we conducted two preregistered read-only audits without replacing any frozen scientific checkpoint.
First, the five matched Progress-10 and Baseline checkpoints were re-evaluated on a fresh common-random-number pool of 500 paired seeds using both the parent-relative estimand in Eq.~\eqref{eq:seed_effect} and direct Progress-versus-Baseline evaluation.
Second, checkpoint-selection stability was assessed on another disjoint 500-pair pool among the three confirmation-eligible candidates from each trajectory, using joint paired-seed resampling and five-fold cross-fitting.
These audits diagnose replication and selection stability without redefining the frozen scientific checkpoints.

We further conduct a post-hoc fixed-checkpoint horizon diagnostic over all ten matched lineages.
For every training seed $s$ and iteration $t\in\{0,\ldots,7\}$, a fresh common-random-number pool of 200 paired team-swap seeds estimates the direct effect
\begin{equation}
D_{s,t}
=
\operatorname{MeanDelta}(P_{s,t},B_{s,t}).
\label{eq:horizon_effect}
\end{equation}
We summarize the mean trajectory, the late contrast $L_s=D_{s,7}-D_{s,4}$, and a per-seed linear slope over iterations 4--7.
This analysis is diagnostic rather than confirmatory and does not select or replace any checkpoint.

Finally, an exploratory five-seed scale/shuffle ablation separates the conventional Baseline ($B_0$), a scale-matched Baseline using the Progress-10 divisor ($B_S$), a semantically shuffled progress control at that scale ($P_S$), and the true Progress-10 treatment ($P_T$).
Its contrasts are reported as mechanism diagnostics with exploratory seed-level intervals, not as equivalence tests or replacements for the preregistered method-level replication.

\subsection{Diagnostics and Statistical Reporting}
\label{sec:diagnostics}
\label{sec:statistics}

We monitor optimization health through Q-value summaries, within-state action-score dispersion, round-target scale, parent divergence, illegal-action checks, and predictor out-of-support rates.
These diagnostics help interpret training instability but do not replace complete-match paired evaluation as the primary measure of playing strength.
For a fixed checkpoint comparison, we report positive/split/negative paired counts, MeanDelta with a bootstrap confidence interval, and raw match win/loss counts.
For the method-level comparison, the independent unit is the training seed rather than the individual evaluation match, so cohort-specific and pooled ten-seed results are reported separately from the uncertainty of a single selected checkpoint.

\section{Results}
\label{sec:results}

\subsection{Predictive Meaning: State--Gain Interactions Improve Held-Out Fit}
\label{sec:results_r2}

Table~\ref{tab:r2_results} compares the constant-gain and state-dependent specifications on 9,568 team-relative boundaries from 390 held-out matches.
Adding pre-round state--gain interactions yields a small but statistically supported improvement in held-out log loss: the match-clustered bootstrap change (M1 minus M0) is $-0.00476$, with a 95\% confidence interval of $[-0.00766,-0.00201]$.

\begin{table}[!t]
\caption{Held-out incremental predictive-value test. M0 uses a constant gain effect and M1 adds state--gain interactions. Lower is better for log loss and Brier score; higher is better for AUROC.}
\label{tab:r2_results}
\centering
\begin{small}
\begin{sc}
\begin{tabular}{lccc}
\toprule
Model & Log loss & Brier & AUROC \\
\midrule
M0 & 0.54458 & 0.18431 & 0.79633 \\
M1 & \textbf{0.53983} & 0.18370 & 0.79572 \\
\bottomrule
\end{tabular}
\end{sc}
\end{small}
\end{table}

The secondary Brier and AUROC comparisons do not show corresponding reliable improvements.
We therefore use \emph{predictively meaningful} in a deliberately narrow sense: the state--gain interaction adds held-out information under log loss, not that it improves every predictive metric.

\begin{figure}[!t]
\centering
\includegraphics[width=\columnwidth]{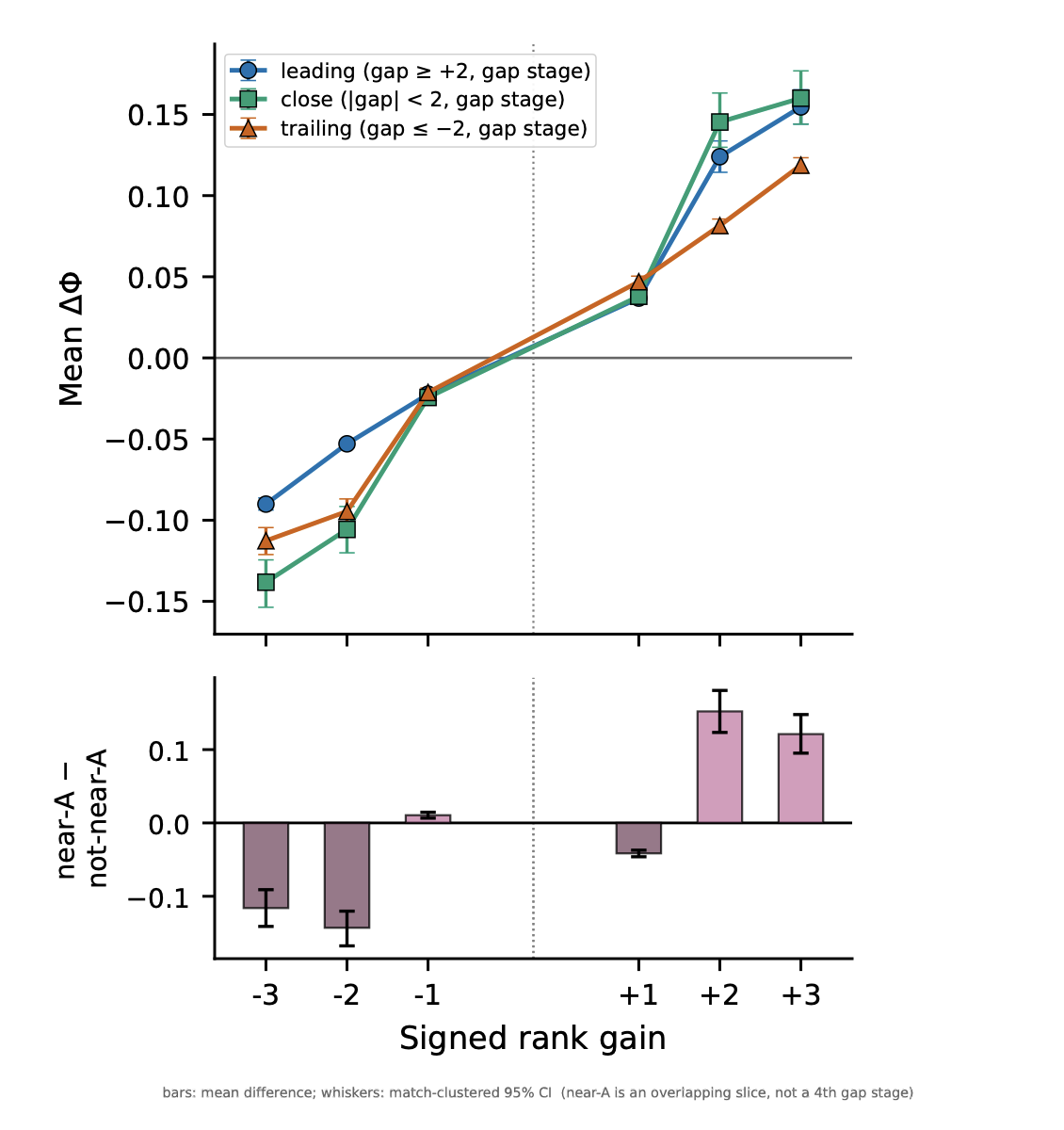}
\caption{Mean $\Delta\Phi$ vs.\ signed rank gain by gap stage (top) and near-A contrast (bottom; overlapping slice, not a fourth exclusive stage).
Error bars: match-clustered 95\% bootstrap CIs.
Stage dependence is strongest at larger $|$gain$|$.}
\label{fig:state_progress}
\end{figure}

Figure~\ref{fig:state_progress} summarizes mean $\Delta\Phi$ by signed gain and gap stage, together with the overlapping Near-A contrast.
The direction is semantically consistent: every negative-gain stratum has negative mean progress and every positive-gain stratum has positive mean progress.
The magnitude is state dependent: 12 of the 18 pairwise gap-stage contrasts have match-clustered 95\% intervals excluding zero, with separation concentrated in larger round outcomes ($|\Delta_k|\geq2$) rather than appearing as a uniform offset.
All six Near-A contrasts also exclude zero, but their direction depends on gain size and the current gap stage.
Empirical match outcomes retain the same stage ordering at fixed large gains, reinforcing that the predictive value of a round outcome depends on the pre-round state.

\subsection{Measurement Check: The Frozen Predictor Tracks Match Outcomes}
\label{sec:results_phi}

On the held-out team-0 view ($n=4{,}784$ boundaries), the calibrated public-state predictor obtains log loss 0.59497, Brier score 0.20462, AUROC 0.75064, and expected calibration error 0.05064 (Table~\ref{tab:phi_results}).

\begin{table}[!t]
\caption{Frozen $\Phi$ performance on the untouched test split.}
\label{tab:phi_results}
\centering
\begin{small}
\begin{sc}
\begin{tabular}{lc}
\toprule
Metric & Test value \\
\midrule
Log loss & 0.59497 \\
Brier score & 0.20462 \\
AUROC & 0.75064 \\
ECE & 0.05064 \\
Clipping rate & 0.000209 \\
Fallback rate & 0 \\
\bottomrule
\end{tabular}
\end{sc}
\end{small}
\end{table}

\begin{figure}[!t]
\centering
\includegraphics[width=0.82\columnwidth]{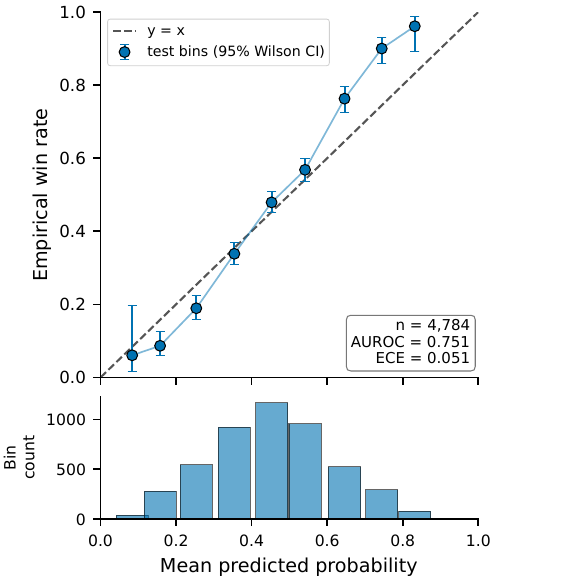}
\caption{Reliability diagram for frozen $\Phi$ on the test split ($n = 4{,}784$); 95\% Wilson intervals per bin, with bin counts below.
Predictions track the diagonal except in sparse upper bins (AUROC $=0.751$, ECE $=0.051$).
Platt scaling was fit on validation only.}
\label{fig:phi_calibration}
\end{figure}

The reliability curve in Figure~\ref{fig:phi_calibration} shows the largest deviations in sparse upper-probability bins.
Opponent-view probabilities are defined as complements, so team-symmetry error is zero by construction.

\subsection{Optimization Probe: Initial Scale Selection}
\label{sec:results_scale}

Table~\ref{tab:scale_selection} summarizes the initial five-seed optimization comparison used to select the progress scale before accessing the final holdout.
On the confirmation protocol, the aggregate parent-relative MeanDelta was approximately $0.120$ for Baseline and $0.154$ for Progress-10.
The corresponding Progress-minus-Baseline aggregate effect was $+0.034$ for Progress-10 and only $+0.004$ for Progress-20.

\begin{table}[!t]
\caption{Five-seed progress-scale selection before the final holdout.
``Agg.\ $\Delta$'' is the Progress-minus-Baseline MeanDelta on the confirmation protocol and is therefore $0$ for Baseline by definition.
Parent-relative aggregates are reported in the text ($0.120$ for Baseline; $0.154$ for Progress-10).
Final marks admission to the locked holdout.}
\label{tab:scale_selection}
\centering
\begin{small}
\begin{sc}
\setlength{\tabcolsep}{3.5pt}
\begin{tabular}{@{}lrrccc@{}}
\toprule
Arm & $\beta$ & $c_a$ & Pos. & Agg.\ $\Delta$ & Final \\
\midrule
Baseline & $0$ & $2.556$ & --- & $0$ & no \\
P10 & $2.002$ & $2.747$ & $3/5$ & $+0.034$ & yes \\
P20 & $4.005$ & $2.948$ & $2/5$ & $+0.004$ & no \\
\bottomrule
\end{tabular}
\end{sc}
\end{small}
\end{table}

Progress-10 achieved the larger aggregate Progress-minus-Baseline improvement and was positive for three of five matched training seeds, so it was selected before the final holdout under the preregistered rule.

\subsection{Optimization Reliability Breaks Across Training Seeds}
\label{sec:results_seeds}

Predictive meaning would translate into optimization reliability only if the selected Progress-10 scale yielded a consistently positive method-level effect across independent training seeds.
In the original confirmation analysis, three of five Progress-10 seed-level effects were positive, but their mean was only $0.034$ with a 95\% bootstrap interval of $[-0.080,0.147]$.
Thus, the five training seeds already exhibit clear heterogeneity and do not establish a positive expected method-level effect, despite the independently verified predictive content of the injected signal.

The preregistered replication is less favorable.
Across seeds 6--10, none of the five Progress-10 effects is positive; the mean is $-0.056$ with a Student-$t$ 95\% confidence interval of $[-0.083,-0.030]$.
This batch therefore rejects a positive mean effect and excludes an improvement of at least $+0.05$ MeanDelta under the preregistered practical-effect criterion.
As a secondary analysis, pooling all ten training seeds gives mean effect $-0.030$ with 95\% confidence interval $[-0.084,0.025]$.
The pooled interval contains zero, so the combined evidence establishes neither a positive nor a negative population-average effect; importantly, it provides no support for a reliable positive method-level gain.
This is the paper's central predictive--optimization gap: held-out predictive value survives its own test, while a reproducible positive policy effect does not survive method-level replication.

\begin{table}[!t]
\caption{Method-level Progress-10 effect across training-seed cohorts. The original interval is a seed bootstrap; the preregistered replication and pooled intervals use Student-$t$ inference over training seeds. Positive counts are descriptive.}
\label{tab:method_replication}
\centering
\begin{small}
\begin{sc}
\setlength{\tabcolsep}{4pt}
\begin{tabular}{@{}lccc@{}}
\toprule
Cohort & Mean $D_s$ & 95\% CI & Positive \\
\midrule
Seeds 1--5 & $+0.034$ & $[-0.080,0.147]$ & $3/5$ \\
Seeds 6--10 & $-0.056$ & $[-0.083,-0.030]$ & $0/5$ \\
Seeds 1--10 & $-0.030$ & $[-0.084,0.025]$ & $3/10$ \\
\bottomrule
\end{tabular}
\end{sc}
\end{small}
\end{table}

As shown in Table~\ref{tab:seed_robustness}, a fresh 500-pair replication preserves the original sign pattern for four of five seeds, and direct Progress-versus-Baseline evaluation confirms both positive and negative lineages.
Cross-fitted selection among confirmation-eligible candidates preserves the negative effects for seeds 3 and 5, so checkpoint selection does not explain the overall positive-versus-negative structure.

\begin{table}[t]
\caption{Seed-level Progress-10 effects under the original confirmation analysis, independent 500-pair replication, and cross-fitted checkpoint-selection audit. ``Parent-rel.'' is the difference between Progress-versus-Parent and Baseline-versus-Parent MeanDelta; ``Direct'' evaluates Progress-10 against its matched Baseline. Cross-fit values are held-out estimates of the original three-candidate selection rule and do not replace the frozen scientific checkpoints.}
\label{tab:seed_robustness}
\centering
\begin{small}
\begin{sc}
\setlength{\tabcolsep}{3.5pt}
\begin{tabular}{@{}crrrr@{}}
\toprule
Seed &
Original &
Parent-rel. &
Direct &
Cross-fit \\
\midrule
1 & $+0.145$ & $+0.128$ & $+0.128$ & $+0.14$ \\
2 & $+0.135$ & $+0.182$ & $+0.074$ & $+0.02$ \\
3 & $-0.125$ & $-0.102$ & $-0.160$ & $-0.13$ \\
4 & $+0.155$ & $-0.044$ & $-0.004$ & $-0.05$ \\
5 & $-0.140$ & $-0.038$ & $-0.068$ & $-0.05$ \\
\bottomrule
\end{tabular}
\end{sc}
\end{small}
\end{table}

We additionally audited selection stability among the three confirmation-eligible checkpoints from each trajectory.
Most selections were stable, with Progress-10 seeds 3--4 and Baseline seed 2 less so; five-fold cross-fitting nonetheless preserved the negative Progress-10 effects for seeds 3 and 5.

\subsection{A Successful Checkpoint Does Not Establish Reliability}
\label{sec:results_final}

Against this heterogeneous ten-seed background, the locked holdout result for the selected Progress-10 scientific checkpoint provides an instructive counterpoint rather than a method-average claim.
This frozen checkpoint is stronger than both the common DMC Parent and its matched Baseline on the locked 1000-pair team-swap evaluation (Table~\ref{tab:final_eval}).
Against the matched Baseline, Progress-10 records 293 double wins, 557 split pairs, and 150 double losses, giving MeanDelta $0.143$ with 95\% confidence interval $[0.103,0.183]$, corresponding to an approximate single-match win rate of $57.2\%$ over the 2{,}000 individual matches.
Against the Parent, MeanDelta is $0.254$ with interval $[0.214,0.294]$ and raw win rate $0.627$.
These results establish that the signal can be beneficial in a particular lineage, but they do not show that the intervention is optimization reliable across training randomness.

\begin{table}[!t]
\caption{Locked paired team-swap evaluation of the selected Progress-10 checkpoint on 1{,}000 pairs. W/S/L are double-win, split, and double-loss counts; MD is MeanDelta.}
\label{tab:final_eval}
\centering
\begin{small}
\begin{sc}
\begin{tabular}{lccc}
\toprule
Comparison & W/S/L & MD & 95\% CI \\
\midrule
vs.\ Parent & 369/516/115 & 0.254 & [0.214, 0.294] \\
vs.\ Baseline & 293/557/150 & 0.143 & [0.103, 0.183] \\
\bottomrule
\end{tabular}
\end{sc}
\end{small}
\end{table}

\subsection{The Gap Is Not Explained by Selection or Training Horizon}
\label{sec:results_diagnostics}

\begin{figure*}[!t]
\centering
\includegraphics[width=0.72\textwidth]{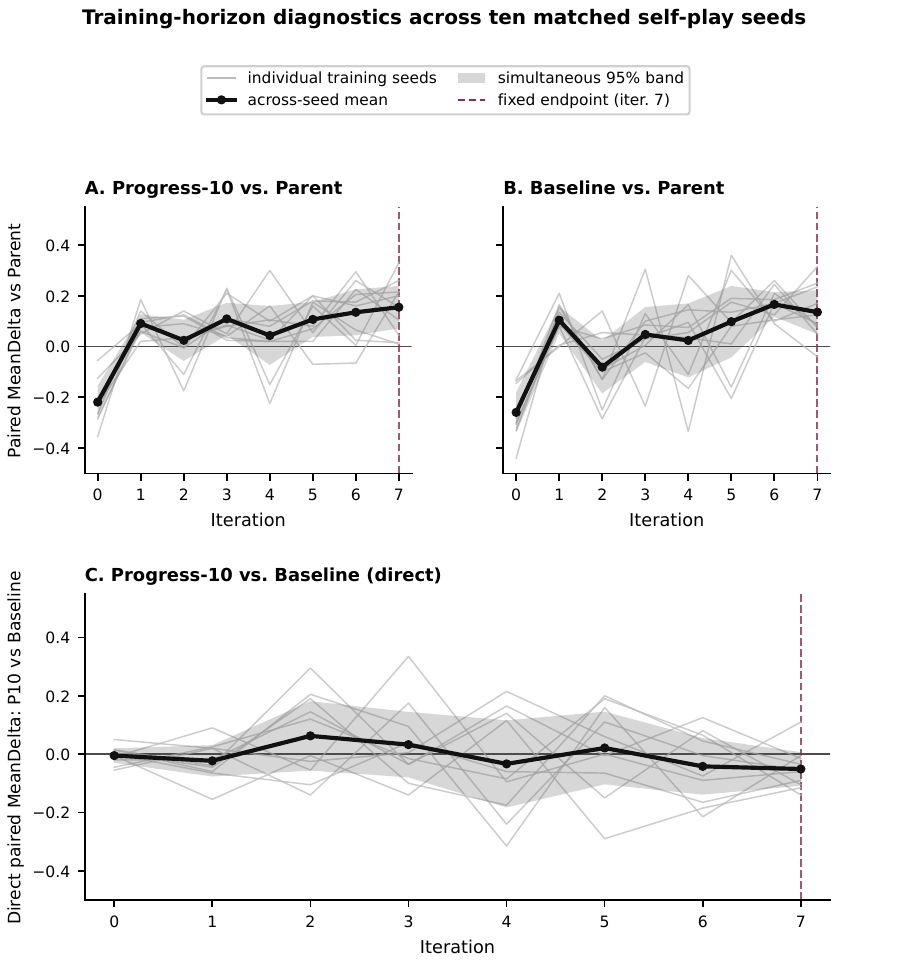}
\caption{Training-horizon diagnostics across ten matched self-play seeds.
Panels A and B show the paired MeanDelta of Progress-10 and Baseline, respectively, against the frozen Parent; Panel C reports direct Progress-10-versus-Baseline evaluation and is therefore the relevant treatment-effect trajectory.
Each matchup uses 200 paired team-swap seeds from the diagnostic common-random-number pool.
Thin gray lines denote individual training seeds, thick black lines denote across-seed means, gray regions show simultaneous 95\% confidence bands over the eight inspected iterations, and the dashed line marks the fixed endpoint at iteration 7.
Both policies improve in absolute strength relative to Parent, whereas their direct comparison fluctuates around zero and shows no common late-stage Progress advantage.
This post-hoc diagnostic is not used for checkpoint selection.}
\label{fig:horizon_diagnostics}
\end{figure*}

Post-hoc audits did not identify a simple measured mechanism separating positive and negative Progress-10 lineages.
Frozen-predictor log loss and Brier score were similar across the groups, and the realized progress-component scale remained close to its nominal 10\% contribution.
Coarse Q dispersion and top-action margins were also similar.
On a fixed 2,000-state bank, matched Progress-10 and Baseline policies retained Spearman action-ranking correlation of approximately $0.94$--$0.95$, while controlled $K=8$ branch rollouts with Parent continuation found no systematic advantage for the action disagreements of positive lineages; the corresponding confidence intervals included zero.
An offline per-iteration analysis likewise found only isolated differences in top-action divergence, with no persistent consecutive-iteration separation satisfying the preregistered gate.
Together with the selection audit, these null results indicate that evaluation noise, checkpoint selection, predictor drift, reward-scale drift, and the measured one-step ranking statistics do not individually explain the seed pattern; they do not exclude unmeasured multi-step self-play interactions.
The predictive--optimization gap therefore cannot be reduced to one of these measured artifacts.

The all-checkpoint horizon diagnostic in Figure~\ref{fig:horizon_diagnostics} covers all ten matched Progress-10--Baseline training-seed pairs and does not support insufficient training duration as an explanation for the missing Progress-specific effect.
Across these ten pairs, the direct Progress-minus-Baseline trajectories repeatedly cross zero, their peaks occur at different iterations, and intermediate gains are generally not retained at the fixed endpoint.
The mean iteration-4-to-7 contrast is $-0.0175$ (95\% CI $[-0.1160,0.0925]$), and the mean late-stage slope is $-0.0116$ per iteration ($[-0.0341,0.0152]$).
Although the iteration-7 pointwise interval is negative, its simultaneous band over the eight inspected horizons still includes zero.
Accordingly, the curves should not be interpreted as proving monotonic deterioration; they show substantial lineage-dependent fluctuation and no common late-stage tendency for Progress-10 to overtake matched Baseline training.

The absence of a reliable relative benefit is compatible with continued absolute learning.
The upper panels show that both Progress-10 and Baseline improve relative to the frozen Parent after early fluctuations, but the direct comparison in Panel C shows no corresponding late-stage Progress-specific improvement.
Longer training may therefore increase absolute policy strength without resolving the absence of a reliable Progress-specific treatment effect.

The exploratory scale/shuffle ablation also fails to isolate a stable semantic benefit.
The true-versus-shuffled progress contrast is $P_T-P_S=-0.032$ with 95\% interval $[-0.182,0.117]$; the scale-only contrast is $B_S-B_0=+0.036$ with interval $[-0.119,0.190]$.
The shuffled-noise contrast $P_S-B_S=-0.002$ and the same-scale true-progress contrast $P_T-B_S=-0.006$ are both near zero.
This separate scale/shuffle ablation uses five seeds per condition and has wide intervals; it detects no stable semantic, scale, or noise advantage and does not establish statistical equivalence among the four conditions.

\section{Discussion}
\label{sec:discussion}

\subsection{Predictive Meaning Is Not Optimization Reliability}

The main result is an asymmetry between two kinds of evidence.
State--gain interactions improve held-out log loss, the frozen predictor tracks eventual match outcomes, and its progress estimates vary coherently with public match context.
These results establish predictive meaning under the stated measurement criteria.
In contrast, the preregistered replication mean is negative, the pooled ten-seed interval includes zero, and relative training trajectories repeatedly change sign.
The same signal therefore fails the stronger criterion of a reproducibly positive optimization effect under the studied DMC protocol.

The successful seed-1 checkpoint is part of this finding rather than an exception to discard.
It shows that progress augmentation can produce a substantially stronger policy in one lineage, ruling out the simple interpretation that the signal is uniformly harmful or inert.
At the same time, its coexistence with negative and null lineages demonstrates why selected-checkpoint efficacy and method-level reliability answer different questions.
The former establishes possibility conditional on a realized training path; the latter asks whether a new training run should be expected to benefit.
We therefore interpret the locked result as \emph{selection-conditioned efficacy}, not as an unconditional estimate of the training procedure's expected effect.

This gap is plausible because prediction and optimization act on different objects.
The predictive test evaluates whether a feature improves outcome estimates under held-out samples from a fixed data-generating distribution, whereas downstream decision quality depends on how prediction errors interact with the optimization objective \cite{smart_predict_optimize,decision_focused}.
The optimization intervention changes finite-sample residuals for executed actions, which changes the policy, which in turn changes subsequent self-play data.
This policy--data feedback is structurally related to performative learning, where deployed decisions alter later data or dynamics \cite{performative_prediction,performative_rl}.
Round-broadcast supervision, approximation error, and stochastic trajectory collection can therefore transform a valid descriptive signal into lineage-dependent updates.
Our diagnostics do not identify a single causal mechanism for that transformation, but they exclude several simpler explanations and show that training longer does not produce a common late-stage advantage.

The practical implication is a two-stage standard for outcome-potential supervision.
Predictor validation should establish that the proposed signal contains the intended information, while independent training-seed replication should establish that the learner converts that information into policy gains reliably.
Neither predictor metrics nor a locked evaluation of one selected checkpoint can replace the second stage, consistent with established concerns about variability and finite-run uncertainty in deep RL evaluation \cite{deep_rl_matters,statistical_precipice}.
This distinction is especially important in online self-play, where optimization changes the distribution on which later supervision is generated.

\subsection{Scope and Limitations}

The conclusion is deliberately bounded to the present construction.
First, $\Phi_\psi$ is predictive rather than causal, so $\Delta\Phi$ should not be interpreted as action-level credit.
Second, the augmentation has no classical PBRS guarantee, and every executed action within a round shares the same label.
Third, the frozen teacher may face distribution shift under later self-play policies.
Fourth, ten Baseline--Progress seed pairs remain a modest sample for characterizing a broad training-run population, and the completed diagnostics do not identify which random source determines whether a lineage benefits.
Finally, the experiment covers one game and one round-broadcast, executed-action DMC learner; it does not establish that predictive meaning and optimization reliability will diverge under policy-gradient, counterfactual, or policy-invariant constructions.

\section{Conclusion}
\label{sec:conclusion}

We used match-progress supervision in GuanDan as a controlled case study of a commonly implicit inference: that an outcome-predictive intermediate signal should also be a reliable optimization signal.
The evidence separates these claims.
Public match context adds held-out predictive value, and progress augmentation can yield a strong selected checkpoint, yet its policy effect is heterogeneous across ten training seeds and shows no common late-stage advantage.
Thus, under the studied self-play protocol, predictive meaning does not imply optimization reliability.
Determining which credit-assignment or policy-optimization mechanisms can convert valid predictive information into reproducible gains is the next question, not an assumption that predictor quality has already answered.


\IEEEtriggeratref{18}
\bibliographystyle{IEEEtran}
\bibliography{guandan}
\end{document}